\documentclass[twoside]{article}
\usepackage[accepted]{aistats2017_no}
\usepackage{times}
\usepackage{epsfig}
\usepackage{graphicx}
\usepackage{hyperref} 
\usepackage{amsmath}
\usepackage{booktabs}  
\usepackage{amssymb}
\usepackage{enumerate}
\usepackage{amsfonts,lscape,color,url}
\usepackage{adjustbox}
\usepackage{graphicx,graphics,wrapfig,epstopdf,subcaption}
\usepackage{algorithm}
\usepackage{algorithmic}
\usepackage{natbib}
\newcommand{\beq}{\begin{equation}}
\newcommand{\eeq}{\end{equation}}
\newcommand{\beqs}{\begin{eqnarray}}
\newcommand{\eeqs}{\end{eqnarray}}
\newcommand{\barr}{\begin{array}}
	\newcommand{\earr}{\end{array}}

\newcommand{\ie}{\textit{i.e.}}

\newcommand{\Imat}{{\bf I}}

\newcommand{\xv}{\boldsymbol{x}}

\newcommand{\zv}{\boldsymbol{z}}
\newcommand{\zerov}{\boldsymbol{0}}

\newcommand{\epsilonv}{{\boldsymbol \epsilon}}

\newcommand{\thetav}{{\boldsymbol \theta}}

\newcommand{\phiv}{{\boldsymbol \phi}}

\newcommand{\psiv}{{\boldsymbol \psi}}

\begin{document}
	
	%
	
	%
	
	\twocolumn[
	
	\aistatstitle{Symmetric Variational Autoencoder and Connections to Adversarial Learning}
	%
	%
	\aistatsauthor{ Liqun Chen \And Shuyang Dai \And  Yunchen Pu \And Chunyuan Li \And Qinliang Su \And Lawrence Carin }
	
	\aistatsaddress{ 	Department of Electrical and Computer Engineering, Duke University \\ \texttt{\{liqun.chen, shuyang.dai, yunchen.pu, chunyuan.li, qinliang.su, lcarin\}@duke.edu}}]
	
	\begin{abstract}
		A new form of the variational autoencoder (VAE) is proposed, based on the {\em symmetric} Kullback-Leibler divergence. It is demonstrated that learning of the resulting symmetric VAE (sVAE) has close connections to previously developed adversarial-learning methods. This relationship helps unify the previously distinct techniques of VAE and adversarially learning, and provides insights that allow us to ameliorate shortcomings with some previously developed adversarial methods. In addition to an analysis that motivates and explains the sVAE, an extensive set of experiments validate the utility of the approach.
	\end{abstract}
	
	\section{Introduction}
	
	Generative models that are {\em descriptive} of data have been widely employed in statistics and machine learning. Factor models (FMs) represent one commonly used generative model \citep{Tipping99probabilisticprincipal}, and mixtures of FMs have been employed to account for more-general data distributions \citep{Ghahramani97theem}. These models typically have latent variables (e.g., factor scores) that are inferred given observed data; the latent variables are often used for a down-stream goal, such as classification \citep{West}. After training, such models are useful for inference tasks given subsequent observed data. However, when one draws from such models, by drawing latent variables from the prior and pushing them through the model to synthesize data, the synthetic data typically do not appear to be realistic. This suggests that while these models may be useful for analyzing observed data in terms of inferred latent variables, they are also capable of describing a large set of data that do not appear to be real.
	
	The generative adversarial network (GAN) \citep{gan} represents a significant recent advance toward development of generative models that are capable of synthesizing realistic data. Such models also employ latent variables, drawn from a simple distribution analogous to the aforementioned prior, and these random variables are fed through a (deep) neural network. The neural network acts as a functional transformation of the original random variables, yielding a model capable of representing sophisticated distributions. Adversarial learning discourages the network from yielding synthetic data that are unrealistic, from the perspective of a learned neural-network-based classifier. However, GANs are notoriously difficult to train, and multiple generalizations and techniques have been developed to improve learning performance \citep{improvegan}, for example Wasserstein GAN (WGAN) \citep{wgan0,WGAN} and energy-based GAN (EB-GAN) \citep{ebgan}.
	
	While the original GAN and variants were capable of synthesizing highly realistic data (e.g., images), the models lacked the ability to infer the latent variables given observed data. This limitation has been mitigated recently by methods like adversarial learned inference (ALI) \citep{ALI}, and related approaches. However, ALI appears to be inadequate from the standpoint of inference, in that, given observed data and associated inferred latent variables, the subsequently synthesized data often do not look particularly close to the original data.
	
	The variational autoencoder (VAE) \citep{vae} is a class of generative models that precedes GAN. VAE learning is based on optimizing a variational lower bound, connected to inferring an approximate posterior distribution on latent variables; such learning is typically not performed in an adversarial manner. VAEs have been demonstrated to be effective models for inferring latent variables, in that the reconstructed data do typically look like the original data, albeit in a blurry manner \citep{ALI}. The form of the VAE has been generalized recently, in terms of the adversarial variational Bayesian (AVB) framework \citep{AVB}. This model yields general forms of encoders and decoders, but it is based on the original variational Bayesian (VB) formulation. 
	The original VB framework yields a lower bound on the log likelihood of the observed data, and therefore model learning is connected to maximum-likelihood (ML) approaches. From the perspective of designing generative models, it has been recognized recently that ML-based learning has limitations \citep{wgan0}: such learning tends to yield models that match observed data, but also have a high probability of generating unrealistic synthetic data. 
	
	The original VAE employs the Kullback-Leibler divergence to constitute the variational lower bound. As is well known, the KL distance metric is asymmetric. We demonstrate that this asymmetry encourages design of decoders (generators) that often yield unrealistic synthetic data when the latent variables are drawn from the prior. From a different but related perspective, the encoder infers latent variables (across all training data) that only encompass a subset of the prior. As demonstrated below, these limitations of the encoder and decoder within conventional VAE learning are intertwined. 
	
	We consequently propose a new {\em symmetric} VAE (sVAE), based on a symmetric form of the KL divergence and associated variational bound. The proposed sVAE is learned using an approach related to that employed in the AVB \citep{AVB}, but in a new manner connected to the symmetric variational bound. Analysis of the sVAE demonstrates that it has close connections to ALI \citep{ALI}, WGAN \citep{WGAN} and to the original GAN \citep{gan} framework; in fact, ALI is recovered exactly, as a special case of the proposed sVAE. This provides a new and explicit linkage between the VAE (after it is made symmetric) and a wide class of adversarially trained generative models. Additionally, with this insight, we are able to ameliorate much of the aforementioned limitations of ALI, from the perspective of data reconstruction. In addition to analyzing properties of the sVAE, we demonstrate excellent performance on an extensive set of experiments.
	
	\section{Review of Variational Autoencoder}
	\subsection{Background}
	
	Assume observed data samples $\xv\sim q(\xv)$, where $q(\xv)$ is the true and unknown distribution we wish to approximate. Consider $p_{\thetav} (\xv|\zv)$, a model with parameters $\thetav$ and latent code $\zv$. With prior $p(\zv)$ on the codes, the modeled generative process is $\xv\sim p_\theta(\xv|\zv)$, with $\zv\sim p(\zv)$. We may marginalize out the latent codes, and hence the model is $\xv\sim p_{\thetav}(\xv)=\int d\zv p_{\thetav}(\xv|\zv)p(\zv)$. To learn $\thetav$, we typically seek to maximize the expected log likelihood: $\hat{\thetav}=\mbox{argmax}_{\thetav} ~\mathbb{E}_{q(\xv)}\log p_{\thetav}(\xv)$,
	where one typically invokes the approximation $\mathbb{E}_{q(\xv)}\log p_{\thetav}(\xv)\approx \frac{1}{N}\sum_{n=1}^N \log p_{\thetav}(\xv_n)$
	assuming $N$ iid observed samples $\{\xv_n\}_{n=1,N}$. 
	
	It is typically intractable to evaluate $p_{\thetav}(\xv)$ directly, as $\int d\zv p_{\thetav}(\xv|\zv)p(\zv)$ generally doesn't have a closed form. Consequently, a typical approach is to consider a model $q_{\phiv}(\zv|\xv)$ for the posterior of the latent code $\zv$ given observed $\xv$, characterized by parameters $\phiv$. Distribution $q_{\phiv}(\zv|\xv)$ is often termed an {\em encoder}, and $p_{\theta}(\xv|\zv)$ is a {\em decoder} \citep{vae}; both are here stochastic, {\em vis-\`a-vis} their deterministic counterparts associated with a traditional autoencoder \citep{AE}. 
	Consider the variational expression
	\beq
	\mathcal{L}_x(\thetav,\phiv)=\mathbb{E}_{q(\xv)}\mathbb{E}_{q_{\phiv}(\zv|\xv)} \log\big[\frac{p_{\thetav}(\xv|\zv)p(\zv)}{q_{\phiv}(\zv|\xv)}\big]\label{eq:VB}
	\eeq
	In practice the expectation wrt $\xv\sim q(\xv)$ is evaluated via sampling, assuming $N$ observed samples $\{\xv_n\}_{n=1,N}$. One typically must also utilize sampling from $q_{\phiv}(\zv|\xv)$ to evaluate the corresponding expectation in (\ref{eq:VB}). Learning is effected as $(\hat{\thetav},\hat{\phiv})=\mbox{argmax}_{\thetav,\phiv} ~\mathcal{L}_x(\thetav,\phiv)$, and a model so learned is termed a variational autoencoder (VAE) \citep{vae}. 
	
	It is well known that $\mathcal{L}_x(\thetav,\phiv)=\mathbb{E}_{q(\xv)}[\log p_{\thetav}(\xv)-\mbox{KL}(q_{\phiv}(\zv|\xv)\|p_{\thetav}(\zv|\xv))]
	\leq\mathbb{E}_{q(\xv)}[\log p_{\thetav}(\xv)]$. 
	Alternatively, the variational expression may be represented as 
	\beq
	\mathcal{L}_x(\thetav,\phiv)
	=-\mbox{KL}(q_{\phiv}(\xv,\zv)\|p_{\thetav}(\xv,\zv))+C_x\label{eq:bound}
	\eeq
	where $q_{\phiv}(\xv,\zv)=q(\xv)q_{\phiv}(\zv|\xv)$, $p_{\thetav}(\xv,\zv)=p(\zv)p_{\thetav}(\xv|\zv)$ and $C_x=\mathbb{E}_{q(\xv)}\log q(\xv)$.
	One may readily show that \begin{small}
		\beqs
		&&\mbox{KL}(q_{\phiv}(\xv,\zv)\|p_{\thetav}(\zv,\zv))\nonumber\\&=&\mathbb{E}_{q(\xv)}\mbox{KL}(q_{\phiv}(\zv|\xv)\|p_{\thetav}(\zv|\xv))+\mbox{KL}(q(\xv)\|p_{\thetav}(\xv))\label{eq:v1}\\
		&=&\mathbb{E}_{q_{\phiv}(\zv)}\mbox{KL}(q_{\phiv}(\xv|\zv)\|p_{\thetav}(\xv|\zv))+\mbox{KL}(q_{\phiv}(\zv)\|p(\zv))\label{eq:v2}
		\eeqs\end{small}
	where $q_{\phiv}(\zv)=\int q(\xv)q_{\phiv}(\zv|\xv)d\xv$. 
	To maximize $\mathcal{L}_x(\thetav,\phiv)$, we seek minimization of $\mbox{KL}(q_{\phiv}(\xv,\zv)\|p_{\thetav}(\zv,\zv))$.
	Hence, from (\ref{eq:v1}) the goal is to align $p_{\thetav}(\xv)$ with $q(\xv)$, while from (\ref{eq:v2}) the goal is to align $q_{\phiv}(\zv)$ with $p(\zv)$. The other terms seek to match the respective conditional distributions. All of these conditions are implied by minimizing $\mbox{KL}(q_{\phiv}(\xv,\zv)\|p_{\thetav}(\zv,\zv))$. However, the KL divergence is {\em asymmetric}, which yields limitations wrt the learned model.
	
	
	\subsection{Limitations of the VAE\label{sec:limitations}}
	
	The support $\mathcal{S}^{\epsilon}_{p(\zv)}$ of a distribution $p(\zv)$ is defined as the member of the set
	$\{  \tilde{\mathcal{S}}^{\epsilon}_{p(\zv)}: \int_{\tilde{\mathcal{S}}^{\epsilon}_{p(\zv)}}p(\zv)d\zv=1-\epsilon\}$
	with minimum size $\|\tilde{\mathcal{S}}^{\epsilon}_{p(\zv)}\|\triangleq\int_{\tilde{\mathcal{S}}^{\epsilon}_{p(\zv)}}d\zv$. We are typically interested in $\epsilon\rightarrow 0^+$.
	For notational convenience we replace $\mathcal{S}^{\epsilon}_{p(\zv)}$ with $\mathcal{S}_{p(\zv)}$, with the understanding $\epsilon$ is small. 
	We also define $\mathcal{S}_{p(\zv)_{-}}$ as the largest set for which $\int_{\mathcal{S}_{p(\zv)_{-}}}p(\zv)d\zv=\epsilon$, and hence $\int_{\mathcal{S}_{p(\zv)}}p(\zv)d\zv+ \int_{\mathcal{S}_{p(\zv)_{-}}}p(\zv)d\zv=1$. For simplicity of exposition, we assume ${\mathcal{S}_{p(\zv)}}$ and ${\mathcal{S}_{p(\zv)_{-}}}$ are unique; the meaning of the subsequent analysis is unaffected by this assumption.
	
	Consider $-\mbox{KL}(q(\xv)\|p_{\thetav}(\xv))=\mathbb{E}_{q(\xv)}\log p_{\thetav}(\xv)-C_x$, which from (\ref{eq:bound}) and (\ref{eq:v1}) we seek to make large when learning $\thetav$. The following discussion borrows insights from \citep{WGAN}, although that analysis was different, in that it was not placed within the context of the VAE.
	Since $\int_{\mathcal{S}_{q(\xv)_{-}}}q(\xv)\log p_{\thetav}(\xv)d\xv\approx 0$,
	$\mathbb{E}_{q(\xv)}\log p_{\thetav}(\xv)\approx\int_{\mathcal{S}_{q(\xv)}}  q(\xv)\log p_{\thetav}(\xv)d\xv$, and $\mathcal{S}_{q(\xv)} = (\mathcal{S}_{q(\xv)}\cap\mathcal{S}_{p_{\thetav}(\xv)}) \cup (\mathcal{S}_{q(\xv)}\cap\mathcal{S}_{p_{\thetav}(\xv)_{-}})$.  
	If $\mathcal{S}_{q(\xv)}\cap\mathcal{S}_{p_{{\thetav}}(\xv)_{-}}\neq\emptyset$, there is a strong (negative) penalty introduced by $\int_{\mathcal{S}_{q(\xv)}\cap\mathcal{S}_{p_{\thetav}(\xv)_{-}}} q(\xv)\log p_{\thetav}(\xv)d\xv$, and therefore maximization of $\mathbb{E}_{q(\xv)}\log p_{\thetav}(\xv)$ encourages $\mathcal{S}_{q(\xv)}\cap\mathcal{S}_{p_{{\thetav}}(\xv)_{-}}=\emptyset$. By contrast, there is not a substantial penalty to  
	$\mathcal{S}_{q(\xv)_{-}}\cap \mathcal{S}_{p_{{\thetav}}(\xv)}\neq\emptyset$. 
	
	Summarizing these conditions, the goal of maximizing $-\mbox{KL}(q(\xv)\|p_{\thetav}(\xv))$ encourages $\mathcal{S}_{q(\xv)}\subset\mathcal{S}_{p_{\thetav}(\xv)}$. This implies that $p_{\thetav}(\xv)$ can synthesize all $\xv$ that may be drawn from $q(\xv)$, but additionally there is (often) high probability that $p_{\thetav}(\xv)$ will synthesize $\xv$ that will {\em not} be drawn from $q(\xv)$. 
	
	Similarly, $-\mbox{KL}(q_{\phiv}(\zv)\|p(\zv))=h(q_{\phiv}(\zv))+\mathbb{E}_{q_{\phiv}(\zv)}\log p(\zv)$ encourages $\mathcal{S}_{q_{\phiv}(\zv)}\subset\mathcal{S}_{p(\zv)}$, and the commensurate goal of increasing differential entropy $h(q_{\phiv}(\zv))=-\mathbb{E}_{q_{\phiv}(\zv)}\log q_{\phiv}(\zv)$ encourages that $\mathcal{S}_{q_{\phiv}(\zv)}\cap\mathcal{S}_{p(\zv)}$ be as large as possible. 
	
\begin{figure}[tb]
		\centering 
		{ 
			\includegraphics[width=0.3\textwidth]{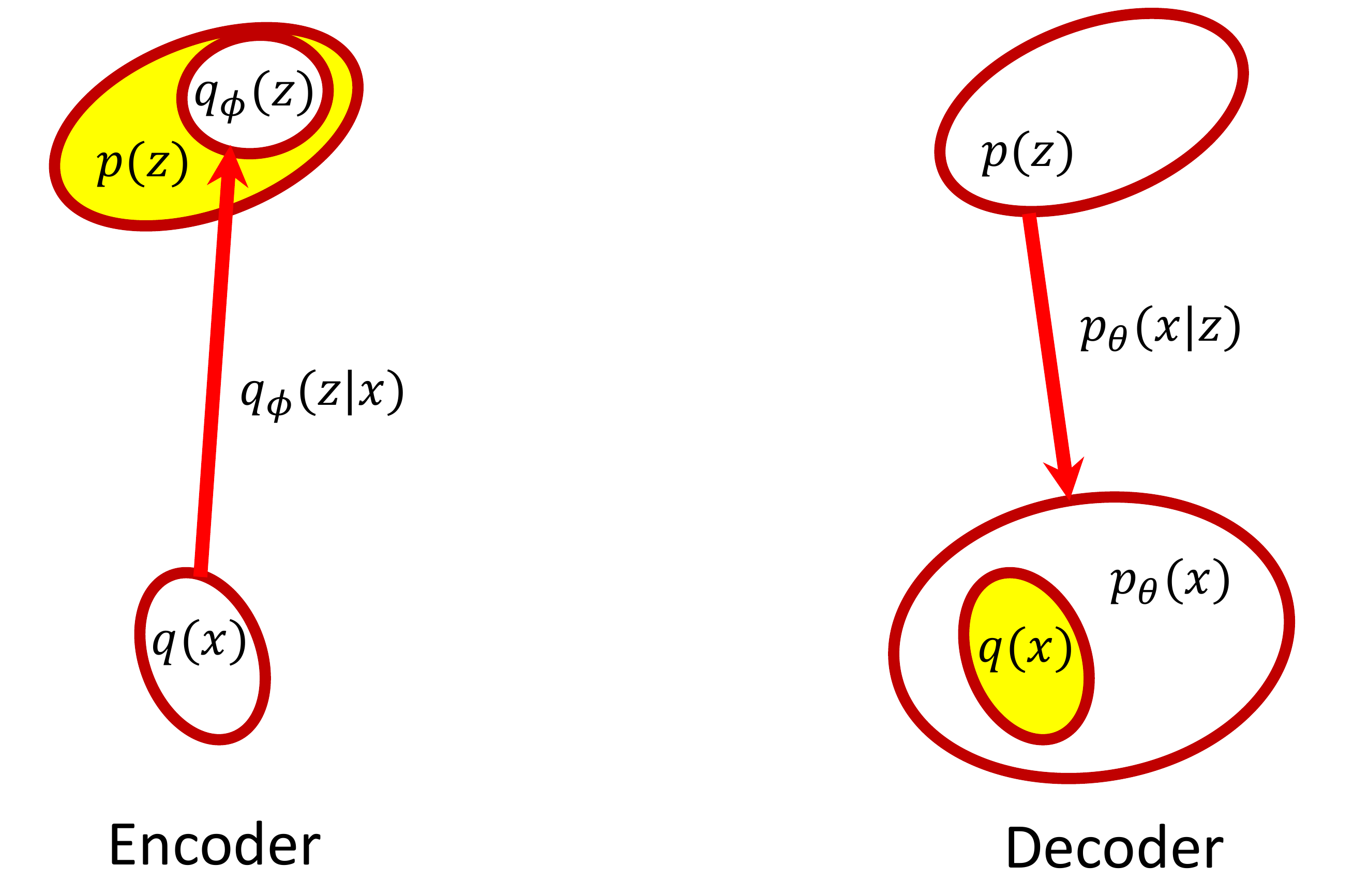}} 
		\caption{\small\label{fig:schematic1}Characteristics of the encoder and decoder of the conventional VAE $\mathcal{L}_x$, for which the support of the distributions satisfy $\mathcal{S}_{q(\xv)}\subset \mathcal{S}_{p_{\thetav}(\xv)}$ and $\mathcal{S}_{q_{\phiv}(\zv)}\subset \mathcal{S}_{p(\zv)}$, implying that the generative model $p_{\thetav}(\xv)$ has a high probability of generating unrealistic draws.\vspace{-4mm}} 		
	\end{figure}

	Hence, the goal of large $-\mbox{KL}(q(\xv)\|p_{\thetav}(\xv))$ and $-\mbox{KL}(q_{\phiv}(\zv)\|p(\zv))$ are saying the same thing, from different perspectives: ($i$) seeking large $-\mbox{KL}(q(\xv)\|p_{\thetav}(\xv))$ implies that there is a high probability that $\xv$ drawn from $p_{\thetav}(\xv)$ will be different from those drawn from $q(\xv)$, and ($ii$) large $-\mbox{KL}(q_{\phiv}(\zv)\|p(\zv))$ implies that $\zv$ drawn from $p(\zv)$ are likely to be different from those drawn from $q_{\phiv}(\zv)$, with $\zv\in \{\mathcal{S}_{p(\zv)}\cap \mathcal{S}_{q_{\phiv}(\zv)_{-}}\}$ responsible for the $\xv$ that are inconsistent with $q(\xv)$. These properties are summarized in Fig. \ref{fig:schematic1}.
	
	\begin{figure}[tb]
		\centering 	{ 
			\includegraphics[width=0.3\textwidth]{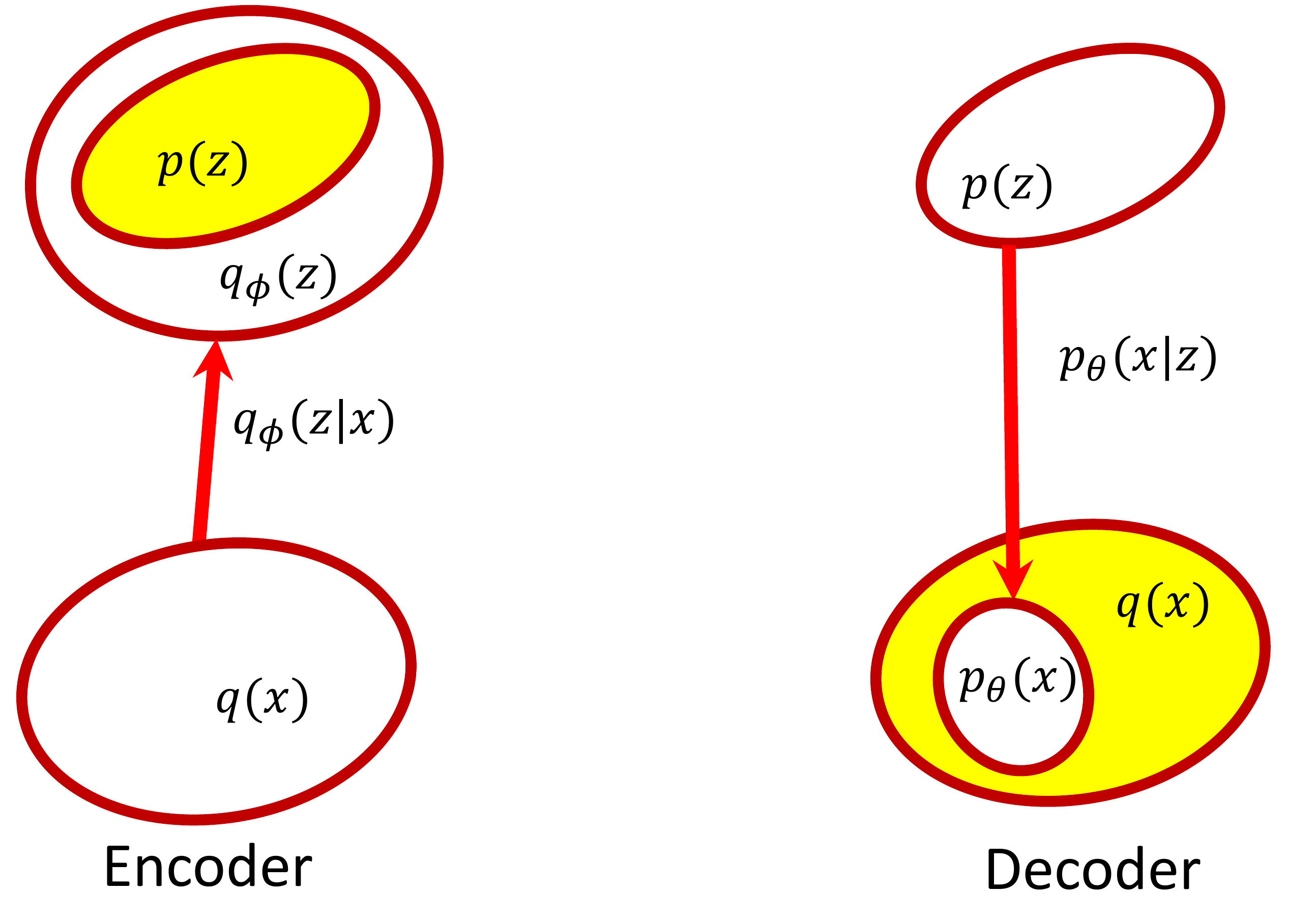}}  	
		\caption{\small\label{fig:schematic2}Characteristics of the new VAE expression, $\mathcal{L}_z$.\vspace{-4mm}} 
		\label{fig:gmm} 
	\end{figure}
	Considering the remaining terms in (\ref{eq:v1}) and (\ref{eq:v2}), and using similar logic on
	$-\mathbb{E}_{q(\xv)}\mbox{KL}(q_{\phiv}(\zv|\xv)\|p_{\thetav}(\zv|\xv))=h(q_{\phiv}(\zv|\xv))+\mathbb{E}_{q(\xv)}\mathbb{E}_{q_{\phiv}(\zv|\xv)}\log p_{\thetav}(\zv|\xv)$, the model encourages $\mathcal{S}_{q_{\phiv}(\zv|\xv)}\subset \mathcal{S}_{p_{\thetav}(\zv|\xv)}$. From $-\mathbb{E}_{q_{\phiv}(\zv)}\mbox{KL}(q_{\phiv}(\xv|\zv)\|p_{\thetav}(\xv|\zv))=h(q_{\phiv}(\xv|\zv))+\mathbb{E}_{q_{\phiv}(\zv)}\mathbb{E}_{q_{\phiv}(\xv|\zv)}\log p_{\thetav}(\xv|\zv)$, the model also encourages $\mathcal{S}_{q_{\phiv}(\xv|\zv)}\subset\mathcal{S}_{p_{\thetav}(\xv|\zv)}$. The differential entropies $h(q_{\phiv}(\zv|\xv))$ and $h(q_{\phiv}(\xv|\zv))$ encourage that $\mathcal{S}_{q_{\phiv}(\zv|\xv)}\cap \mathcal{S}_{p_{\thetav}(\zv|\xv)}$ and $\mathcal{S}_{q_{\phiv}(\xv|\zv)}\cap\mathcal{S}_{p_{\thetav}(\xv|\zv)}$ be as large as possible. Since $\mathcal{S}_{q_{\phiv}(\zv|\xv)}\subset\mathcal{S}_{p_{\thetav}(\zv|\xv)}$, it is anticipated that $q_{\phiv}(\zv|\xv)$ will under-estimate the variance of $p_{\thetav}(\xv|\zv)$, as is common with the variational approximation to the posterior \citep{VB_review}. 

	\section{Refined VAE: Imposition of Symmetry \label{sec:svae}}
	
	\subsection{Symmetric KL divergence}
	
	Consider the new variational expression
	\beqs
	\mathcal{L}_z(\thetav,\phiv)&=&\mathbb{E}_{p(\zv)}\mathbb{E}_{p_{\thetav}(\xv|\zv)} \log\big[\frac{q_{\phiv}(\zv|\xv)q(\xv)}{p_{\thetav}(\xv|\zv)}\big]\\&=&-\mbox{KL}(p_{\thetav}(\xv,\zv)\|q_{\phiv}(\xv,\zv))+C_z
	\eeqs
	where $C_z=-h(p(\zv))$. Using logic analogous to that applied to $\mathcal{L}_x$, maximization of $\mathcal{L}_z$ encourages distribution supports reflected in Fig. \ref{fig:schematic2}.
	
	Defining $\mathcal{L}_{xz}(\thetav,\phiv)=\mathcal{L}_x(\thetav,\phiv)+\mathcal{L}_z(\thetav,\phiv)$, we have
	\beq
	\mathcal{L}_{xz}(\thetav,\phiv)=-\mbox{KL}_s(q_{\phiv}(\xv,\zv)\|p_{\thetav}(\xv,\zv))+K
	\eeq
	where $K=C_x+C_z$, and the {\em symmetric} KL divergence is $\mbox{KL}_s(q_{\phiv}(\xv,\zv)\|p_{\thetav}(\xv,\zv))\triangleq\mbox{KL}(q_{\phiv}(\xv,\zv)\|p_{\thetav}(\xv,\zv))+\mbox{KL}(p_{\thetav}(\xv,\zv)\|q_{\phiv}(\xv,\zv))$. Maximization of $\mathcal{L}_{xz}(\thetav,\phiv)$ seeks minimizing $\mbox{KL}_s(q_{\phiv}(\xv,\zv)\|p_{\thetav}(\xv,\zv))$, which {\em simultaneously} imposes the conditions summarized in Figs. \ref{fig:schematic1} and \ref{fig:schematic2}. 
	
	One may show that
	\begin{small}
		\beqs
		&&\mbox{KL}_s(q_{\phiv}(\xv,\zv)\|p_{\thetav}(\xv,\zv))=\mathbb{E}_{p(\zv)}\mbox{KL}(p_{\thetav}(\xv|\zv)\|q_{\phiv}(\xv|\zv))\nonumber\\&&~+\mathbb{E}_{q_{\phiv}(\zv)}\mbox{KL}(q_{\phiv}(\xv|\zv)\|p_{\thetav}(\xv|\zv))+\mbox{KL}_s(p(\zv)\|q_{\phiv}(\zv))\label{eq:carin1}\\
		&&~~~~~~~~~~~~~~~~~~~~~~~~~~~~~~~~~~~~~~~~~~=\mathbb{E}_{p_{\thetav}(\xv)}\mbox{KL}(p_{\thetav}(\zv|\xv)\|q_{\phiv}(\zv|\xv))\nonumber\\&&~+\mathbb{E}_{q(\xv)}\mbox{KL}(q_{\phiv}(\zv|\xv)\|p_{\thetav}(\zv|\xv))+\mbox{KL}_s(p_{\thetav}(\xv)\|q(\xv))\label{eq:carin2}
		\eeqs
	\end{small}
	
	Considering the representation in (\ref{eq:carin2}), the goal of small $\mbox{KL}_s(p_{\thetav}(\xv)\|q(\xv))$ encourages
	$\mathcal{S}_{q(\xv)}\subset \mathcal{S}_{p_{\thetav}(\xv)}$ {\em and} $\mathcal{S}_{p_{\thetav}(\xv)}\subset\mathcal{S}_{q(\xv)}$, and hence that $\mathcal{S}_{q(\xv)}=\mathcal{S}_{p_{{\thetav}}(\xv)}$. Further, since $-\mbox{KL}_s(p_{\thetav}(\xv)\|q(\xv))=\mathbb{E}_{q(\xv)}\log p_{\thetav}(\xv)+\mathbb{E}_{p_{\thetav}(\xv)}\log q(\xv)+h(p_{\thetav}(\xv))-C_x$, maximization of $-\mbox{KL}_s(p_{\thetav}(\xv)\|q(\xv))$ seeks to minimize the cross-entropy between $q(\xv)$ and $p_{\thetav}(\xv)$, encouraging a complete matching of the distributions $q(\xv)$ and $p_{\thetav}(\xv)$, not just shared support. From (\ref{eq:carin1}), a match is simultaneously encouraged between $p(\zv)$ and $q_{\phiv}(\zv)$. Further, the respective conditional distributions are also encouraged to match.  

	\subsection{Adversarial solution\label{sec:adversarial}}

	Assuming fixed $(\thetav,\phiv)$, and using logic analogous to Proposition 1 in \citep{AVB}, we consider
	\beqs
	g(\psiv)&=&\mathbb{E}_{p_{\phiv}(\xv,\zv)} \log (1-\sigma(f_{\psiv}(\xv,\zv))\nonumber\\ &+&\mathbb{E}_{p_{\thetav}(\xv,\zv)} \log \sigma(f_{\psiv}(\xv,\zv)) \label{eq:g}
	\eeqs
	where $\sigma(\zeta)=1/(1+\exp(-\zeta))$. The scalar function $f_{\psiv}(\xv,\zv)$ is represented by a deep neural network with parameters $\psiv$, and network inputs $(\xv,\zv)$. For fixed $(\thetav,\phiv)$, the parameters $\psiv^*$ that maximize $g(\psiv)$ yield 
	\beq f_{\psiv^*}(\xv,\zv)=\log p_{\thetav}(\xv,\zv)-\log q_{\phiv}(\xv,\zv)\label{eq:match}
	\eeq
	and hence
	\beqs
	\mathcal{L}_x(\thetav,\phiv)&=&\mathbb{E}_{q_{\phiv}(\xv,\zv)} f_{\psiv^*}(\xv,\zv)+C_x\label{eq:x}\\
	\mathcal{L}_z(\thetav,\phiv)&=&-\mathbb{E}_{p_{\thetav}(\xv,\zv)} f_{\psiv^*}(\xv,\zv)+C_z\label{eq:z}
	\eeqs
	Hence, to optimize $\mathcal{L}_{xz}(\thetav,\phiv)$ we consider the cost function
	\beqs
	\ell (\thetav,\phiv;\psiv^*)&=&\mathbb{E}_{q_{\phiv}(\xv,\zv)} f_{\psiv^*}(\xv,\zv)\nonumber\\&&~-\mathbb{E}_{p_{\thetav}(\xv,\zv)} f_{\psiv^*}(\xv,\zv)\label{eq:newALI}
	\eeqs
	Assuming (\ref{eq:match}) holds, we have
	\beq
	\ell (\thetav,\phiv;\psiv^*)=-\mbox{KL}_s(q_{\phiv}(\xv,\zv)\|p_{\thetav}(\xv,\zv))\leq 0
	\eeq
	and the goal is to achieve $\ell (\thetav,\phiv;\psiv^*)=0$ through joint optimization of $(\thetav,\phiv;\psiv^*)$. Model learning consists of alternating between (\ref{eq:g}) and (\ref{eq:newALI}), maximizing (\ref{eq:g}) wrt $\psiv$ with $(\thetav,\phiv)$ fixed, and maximizing (\ref{eq:newALI}) wrt $(\thetav,\phiv)$ with $\psiv$ fixed. 
	
	The expectations in (\ref{eq:g}) and (\ref{eq:newALI}) are approximated by averaging over samples, and therefore to implement this solution we need only be able to sample from $p_{\thetav}(\xv|\zv)$ and $q_{\phiv}(\zv|\xv)$, and we do not require explicit forms for these distributions. For example, a draw from $q_{\phiv}(\zv|\xv)$ may be constituted as $\zv=h_{\phiv}(\xv,\epsilonv)$, where $h_{\phiv}(\xv,\epsilonv)$ is implemented as a neural network with parameters $\phiv$ and $\epsilonv\sim\mathcal{N}(\zerov,\Imat)$.\\
	
	
	\subsection{Interpretation in terms of LRT statistic}
	
	In (\ref{eq:g}) a classifier is designed to distinguish between samples $(\xv,\zv)$ drawn from $p_{\thetav}(\xv,\zv)=p(\zv)p_{\thetav}(\xv|\zv)$ and from $q_{\phiv}(\xv,\zv)=q(\xv)q_{\phiv}(\zv|\xv)$. Implicit in that expression is that there is equal probability that either of these distributions are selected for drawing $(\xv,\zv)$, i.e., that
	$
	(\xv,\zv)\sim [p_{\thetav}(\xv,\zv)+q_{\phiv}(\xv,\zv)]/2
	$. 
	Under this assumption, given observed $(\xv,\zv)$, the probability of it being drawn from $p_{\thetav}(\xv,\zv)$ is $p_{\thetav}(\xv,\zv)/(p_{\thetav}(\xv,\zv)+q_{\phiv}(\xv,\zv))$, and the probability of it being drawn from $q_{\phiv}(\xv,\zv)$ is $q_{\phiv}(\xv,\zv)/(p_{\thetav}(\xv,\zv)+q_{\phiv}(\xv,\zv))$ \citep{gan}. Since the denominator $p_{\thetav}(\xv,\zv)+q_{\phiv}(\xv,\zv)$ is shared by these distributions, and assuming function $p_{\thetav}(\xv,\zv)/q_{\phiv}(\xv,\zv)$ is known, an observed $(\xv,\zv)$ is inferred as being drawn from the underlying distributions as 
	\beqs
	\mbox{if}~p_{\thetav}(\xv,\zv)/q_{\phiv}(\xv,\zv)>1,~~(\xv,\zv)\rightarrow p_{\thetav}(\xv,\zv)\\
	\mbox{if}~p_{\thetav}(\xv,\zv)/q_{\phiv}(\xv,\zv)<1,~~(\xv,\zv)\rightarrow q_{\phiv}(\xv,\zv)
	\eeqs
	This is the well-known likelihood ratio test (LRT) \citep{vanTrees}, and is reflected by (\ref{eq:match}). We have therefore derived a learning procedure based on the log-LRT, as reflected in (\ref{eq:newALI}). The solution is ``adversarial,'' in the sense that when optimizing $(\thetav,\phiv)$ the objective in (\ref{eq:newALI}) seeks to ``fool'' the LRT test statistic, while for fixed $(\thetav,\phiv)$ maximization of (\ref{eq:g}) wrt $\psiv$ corresponds to updating the LRT. This adversarial solution comes as a natural consequence of symmetrizing the traditional VAE learning procedure.
	
	\section{Connections to Prior Work\label{sec:previouswork}}
	
\subsection{Adversarially Learned Inference\label{sec:ALI}}
	
	The adversarially learned inference (ALI) \citep{ALI} framework seeks to learn both an encoder and decoder, like the approach proposed above, and is based on optimizing
	\beqs
	(\hat{\thetav},\hat{\phiv})&=&\mbox{argmin}_{\thetav,\phiv}\max_{\psiv}\{\mathbb{E}_{p_{\thetav}(\xv,\zv)}\log\sigma(f_{\psiv}(\xv,\zv))\nonumber\\&~&+ \mathbb{E}_{q_{\phiv}(\xv,\zv)}\log(1-\sigma(f_{\psiv}(\xv,\zv)))\}\label{eq:ALI}
	\eeqs
	This has similarities to the proposed approach, in that the term $\max_{\psiv}\mathbb{E}_{p_{\thetav}(\xv,\zv)}\log\sigma(f_{\psiv}(\xv,\zv))+ \mathbb{E}_{q_{\phiv}(\xv,\zv)}\log(1-\sigma(f_{\psiv}(\xv,\zv)))$ is identical to our maximization of (\ref{eq:g}) wrt $\psiv$. However, in the proposed approach, rather than directly then optimizing wrt $(\thetav,\phiv)$, as in (\ref{eq:ALI}), in (\ref{eq:newALI}) the result from this term is used to define $f_{\psiv^*}(\xv,\zv)$, which is then employed in (\ref{eq:newALI}) to subsequently optimize over $(\thetav,\phiv)$.
	
	Note that $\log \sigma (\cdot)$ is a monotonically increasing function, and therefore we may replace (\ref{eq:newALI}) as
	\begin{small}
		\beqs
		\ell^\prime (\thetav,\phiv;\psiv^*)&=&\mathbb{E}_{q_{\phiv}(\xv,\zv)} \log\sigma(f_{\psiv^*}(\xv,\zv))\nonumber\\&&+ \mathbb{E}_{p_{\thetav}(\xv,\zv)} \log\sigma(-f_{\psiv^*}(\xv,\zv))\label{eq:GAN}
		\eeqs
	\end{small}
	and note $\sigma(-f_{\psiv^*}(\xv,\zv;\thetav,\phiv))=1-\sigma(f_{\psiv^*}(\xv,\zv;\thetav,\phiv))$. Maximizing (\ref{eq:GAN}) wrt $(\thetav,\phiv)$ with fixed $\psiv^*$ corresponds to the minimization wrt $(\thetav,\phiv)$ reflected in (\ref{eq:ALI}). Hence, the proposed approach is {\em exactly} ALI, if in (\ref{eq:newALI}) we replace $\pm f_{\psiv^*}$ with $\log\sigma(\pm f_{\psiv^*})$.  

\subsection{Original GAN}
	
	The proposed approach assumed both a decoder $p_{\thetav}(\xv|\zv)$ and an encoder $p_{\phiv}(\zv|\xv)$, and we considered the symmetric $\mbox{KL}_s(q_{\phiv}(\xv,\zv)\|p_{\thetav}(\xv,\zv))$. We now simplify the model for the case in which we only have a decoder, and the synthesized data are drawn $\xv\sim p_{\thetav}(\xv|\zv)$ with $\zv\sim p(\zv)$, and we wish to learn $\thetav$ such that data synthesized in this manner match observed data $\xv\sim q(\xv)$. Consider the symmetric
	\beqs
	\mbox{KL}_s(q(\xv)\|p_{\thetav}(\xv))&=&\mathbb{E}_{p(\zv)}\mathbb{E}_{p_{\thetav}(\xv|\zv)}f_{\psiv^*}(\xv)\nonumber\\
	&&-\mathbb{E}_{q(\xv)}f_{\psiv^*}(\xv)
	\eeqs
	where for fixed $\thetav$
	\beq
	f_{\psiv^*}(\xv)=\log(p_{\thetav}(\xv)/q(\xv))\label{eq:LRT}
	\eeq
	We consider a simplified form of (\ref{eq:g}), specifically 
	\beqs
	g(\psiv)&=&\mathbb{E}_{p(\zv)}\mathbb{E}_{p_{\thetav}(\xv|\zv)} \log \sigma(f_{\psiv}(\xv))\nonumber\\&~&+\mathbb{E}_{q(\xv)} \log (1-\sigma(f_{\psiv}(\xv))\label{eq:g1}
	\eeqs
	which we seek to maximize wrt $\psiv$ with fixed $\thetav$, with optimal solution as in (\ref{eq:LRT}). We optimize $\thetav$ seeking to maximize $-\mbox{KL}_s(q(\xv)\|p_{\thetav}(\xv))$, as
	$\mbox{argmax}_{\thetav}~\ell (\thetav;\psiv^*)
	$ where
	\beq
	\ell (\thetav;\psiv^*)
	=\mathbb{E}_{q(\xv)} f_{\psiv^*}(\xv)- \mathbb{E}_{p_{\thetav}(\xv,\zv)} f_{\psiv^*}(\xv) \label{eq:newWGAN}
	\eeq
with $\mathbb{E}_{q(\xv)} f_{\psiv^*}(\xv)$ independent of the update parameter $\thetav$. We observe that in seeking to maximize $\ell (\thetav;\psiv^*)$, parameters $\thetav$ are updated as to ``fool'' the log-LRT $\log [{q(\xv)}/{p_{\thetav}(\xv)}]$. Learning consists of iteratively updating $\psiv$ by maximizing $g(\psiv)$ and updating $\thetav$ by maximizing $\ell(\thetav;\psiv^*)$.
	
	Recall that $\log \sigma (\cdot)$ is a monotonically increasing function, and therefore we may replace (\ref{eq:newWGAN}) as
	\beq
	\ell^\prime (\thetav;\psiv^*)=\mathbb{E}_{p_{\thetav}(\xv,\zv)} \log\sigma(-f_{\psiv^*}(\xv))\label{eq:GAN1}
	\eeq
	Using the same logic as discussed above in the context of ALI, maximizing $\ell^\prime (\thetav;\psiv^*)$ wrt $\thetav$ may be replaced by minimization, by transforming $\sigma(\mu)\rightarrow \sigma(-\mu)$. With this simple modification, minimizing the modified (\ref{eq:GAN1}) wrt $\thetav$ and maximizing (\ref{eq:g1}) wrt $\psiv$, we {\em exactly} recover the original GAN \citep{gan}, for the special (but common) case of a sigmoidal discriminator.
	
\subsection{Wasserstein GAN\label{sec:WGAN}}
	
	The Wasserstein GAN (WGAN) \citep{WGAN} setup is represented as 
	\beq
	{\thetav}=\mbox{argmin}_{\thetav}\max_{\psiv}\{\mathbb{E}_{q(\xv)} f_{\psiv}(\xv)-\mathbb{E}_{p_{\thetav}(\xv,\zv)} f_{\psiv}(\xv)\}\label{eq:wgan}
	\eeq
	where $f_{\psiv}(\xv)$ must be a 1-Lipschitz function. Typically $f_{\psiv}(\xv)$ is represented by a neural network with parameters $\psiv$, with parameter clipping or $\ell_2$ regularization on the weights (to constrain the amplitude of $f_{\psiv}(\xv)$). Note that WGAN is closely related to (\ref{eq:newWGAN}), but in WGAN $f_{\psiv}(\xv)$ doesn't make an explicit connection to the underlying likelihood ratio, as in (\ref{eq:LRT}).
	
	
	It is believed that the current paper is the first to consider symmetric variational learning, introducing $\mathcal{L}_z$, from which we have made explicit connections to previously developed adversarial-learning methods. Previous efforts have been made to match $q_{\phiv}(\zv)$ to $p(\zv)$, which is a consequence of the proposed symmetric VAE (sVAE). For example, \citep{AdversarialAE} introduced a modification to the original VAE formulation, but it loses connection to the variational lower bound \citep{AVB}.
	
\subsection{Amelioration of vanishing gradients}

As discussed in \citep{WGAN}, a key distinction between the WGAN framework in (\ref{eq:wgan}) and the original GAN \citep{gan} is that the latter uses a binary discriminator to distinguish real and synthesized data; the $f_{\psiv}(\xv)$ in WGAN is a 1-Lipschitz function, rather than an explicit discriminator. A challenge with GAN is that as the discriminator gets better at distinguishing real and synthetic data, the gradients wrt the discriminator parameters vanish, and learning is undermined. The WGAN was designed to ameliorate this problem \citep{WGAN}. 

From the discussion in Section \ref{sec:ALI}, we note that the key distinction between the proposed sVAE and ALI is that the latter uses a binary discriminator to distinguish $(\xv,\zv)$ manifested via the generator from $(\xv,\zv)$ manifested via the encoder. By contrast, the sVAE uses a log-LRT, rather than a binary classifier, with it inferred in an adversarial manner. ALI is therefore undermined by vanishing gradients as the binary discriminator gets better, with this avoided by sVAE. The sVAE brings the same intuition associated with WGAN (addressing vanishing gradients) to a generalized VAE framework, with a generator {\em and} a decoder; WGAN only considers a generator. Further, as discussed in Section \ref{sec:WGAN}, unlike WGAN, which requires gradient clipping or other forms of regularization to approximate 1-Lipschitz functions, in the proposed sVAE the $f_{\psiv}(\xv,\zv)$ arises naturally from the symmetrized VAE and we do not require imposition of Lipschitz conditions. As discussed in Section \ref{sec:experiments}, this simplification has yielded robustness in implementation.

	\section{Model Augmentation \label{sec:augmentation}}
	
	A significant limitation of the original ALI setup is an inability to accurately reconstruct observed data via the process $\xv\rightarrow\zv\rightarrow \hat{\xv}$ \citep{ALI}. With the proposed sVAE, which is intimately connected to ALI, we may readily address this shortcoming. The variational expressions discussed above may be written as
	$\mathcal{L}_x=\mathbb{E}_{q_{\phiv}(\xv,\zv)}\log p_{\thetav}(\xv|\zv)-\mathbb{E}_{q(\xv)}\mbox{KL}(q_{\phiv}(\zv|\xv)\|p(\zv))$ and 
	$\mathcal{L}_z=\mathbb{E}_{p_{\thetav}(\xv,\zv)}\log p_{\phiv}(\zv|\xv)-\mathbb{E}_{p(\zv)}\mbox{KL}(p_{\thetav}(\xv|\zv)\|q(\xv))$.
	In both of these expressions, the first term to the right of the equality enforces model fit, and the second term penalizes the posterior distribution {\em for individual data samples} for being dissimilar from the prior (i.e., penalizes $q_{\phiv}(\zv|\xv)$ from being dissimilar from $p(\zv)$, and likewise wrt $p_{\thetav}(\xv|\zv)$ and $q(\xv)$). 
	The proposed sVAE encourages the {\em cumulative} distributions $q_{\phiv}(\zv)$ and $p_{\thetav}(\xv)$ to match $p(\zv)$ and $q(\xv)$, respectively. By simultaneously encouraging more peaked $q_{\phiv}(\zv|\xv)$ and $p_{\thetav}(\xv|\zv)$, we anticipate better ``cycle consistency'' \citep{cycle} and hence more accurate reconstructions.
	
	To encourage $q_{\phiv}(\zv|\xv)$ that are more peaked in the space of $\zv$ for individual $\xv$, and also to consider more peaked $p_{\thetav}(\xv|\zv)$, we may augment the variational expressions as
	\beqs
	\mathcal{L}_x^\prime&=&(\lambda+1)\mathbb{E}_{q_{\phiv}(\xv,\zv)}\log p_{\thetav}(\xv|\zv)\nonumber\\&~&-\mathbb{E}_{q(\xv)}\mbox{KL}(q_{\phiv}(\zv|\xv)\|p(\zv))\label{eq:lx}\\
	\mathcal{L}_z^\prime&=&(\lambda+1)\mathbb{E}_{p_{\thetav}(\xv,\zv)}\log p_{\phiv}(\zv|\xv)\nonumber\\&~&-\mathbb{E}_{p(\zv)}\mbox{KL}(p_{\thetav}(\xv|\zv)\|q(\xv))\label{eq:lz}	\eeqs
	where $\lambda\geq 0$. For $\lambda=0$ the original variational expressions are retained, and for $\lambda>0$, $q_{\phiv}(\zv|\xv)$ and $p_{\thetav}(\xv|\zv)$ are allowed to diverge more from $p(\zv)$ and $q(\xv)$, respectively, while placing more emphasis on the data-fit terms. Defining $\mathcal{L}_{xz}^\prime=\mathcal{L}_x^\prime+\mathcal{L}_z^\prime$, we have
	\beqs 
	\mathcal{L}_{xz}^\prime=\mathcal{L}_{xz}&+&\lambda[\mathbb{E}_{q_{\phiv}(\xv,\zv)}\log p_{\thetav}(\xv|\zv)\nonumber\\&~&~+\mathbb{E}_{p_{\thetav}(\xv,\zv)}\log p_{\phiv}(\zv|\xv)]\label{eq:augment}
	\eeqs
	Model learning is the same as discussed in Sec. \ref{sec:adversarial}, with the modification\begin{small}
		\beqs
		\ell^\prime (\thetav,\phiv;\psiv^*)=&\mathbb{E}_{q_{\phiv}(\xv,\zv)} [f_{\psiv^*}(\xv,\zv)+\lambda\log p_{\thetav}(\xv|\zv)]\nonumber\\-&\mathbb{E}_{p_{\thetav}(\xv,\zv)} [f_{\psiv^*}(\xv,\zv)-\lambda\log p_{\phiv}(\zv|\xv)]\label{eq:svaer}
		\eeqs\end{small}
	%
	%
A disadvantage of this approach is that it requires explicit forms for $p_{\thetav}(\xv|\zv)$ and $p_{\phiv}(\zv|\xv)$, while the setup in Sec. \ref{sec:adversarial} only requires the ability to sample from these distributions.
	
We can now make a connection to additional related work, particularly \citep{Yunchen_NIPS17}, which considered a similar setup to (\ref{eq:lx}) and (\ref{eq:lz}), for the special case of $\lambda=1$. While \citep{Yunchen_NIPS17} had a similar idea of using a symmetrized VAE, they didn't make the theoretical justification presented in Section \ref{sec:svae}. Further, and more importantly, the way in which learning was performed in \citep{Yunchen_NIPS17} is distinct from that applied here, in that \citep{Yunchen_NIPS17} required an additional adversarial learning step, increasing implementation complexity. Consequently, \citep{Yunchen_NIPS17} did not use adversarial learning to approximate the log-LRT, and therefore it cannot make the explicit connection to ALI and WGAN that were made in Sections \ref{sec:ALI} and \ref{sec:WGAN}, respectively.
	
	\section{Experiments\label{sec:experiments}}
	
In addition to evaluating our model on a toy dataset, we consider MNIST, CelebA and CIFAR-10 for both reconstruction and generation tasks. As done for the model ALI with Cross Entropy regularization (ALICE) \citep{lialice}, we also add the augmentation term ($\lambda > 0$ as discussed in Sec. \ref{sec:augmentation}) to sVAE as a regularizer, and denote the new model as sVAE-r. More specifically, we show the results based on the two models:  $i$)
	sVAE: the model is developed in Sec. \ref{sec:svae} to optimize $g(\psiv)$ in (\ref{eq:g}) and $\ell (\thetav,\phiv;\psiv^*)$ in (\ref{eq:newALI}). $ii$) sVAE-r: the model is sVAE with regularization term to optimize $g(\psiv)$ in (\ref{eq:g}) and 
	$\ell^\prime (\thetav,\phiv;\psiv^*)$ in (\ref{eq:svaer}).
	The quantitative evaluation is based on the mean square error (MSE) of reconstructions, log-likelihood calculated via the annealed importance sampling (AIS) \citep{wu2016quantitative}, and inception score (IS) \citep{improvegan}.

	All parameters are initialized with Xavier \citep{xavier} and optimized using Adam \citep{adam} with learning rate of 0.0001. 
	No dataset-specific tuning or regularization, other than dropout \citep{dropout}, is performed. The architectures for the encoder, decoder and discriminator are detailed in the Appendix. All experimental results were performed on a single NVIDIA TITAN X GPU.
	
	
	\begin{figure}[!t]
		\centering 
		{ 
			\includegraphics[width=0.48\textwidth]{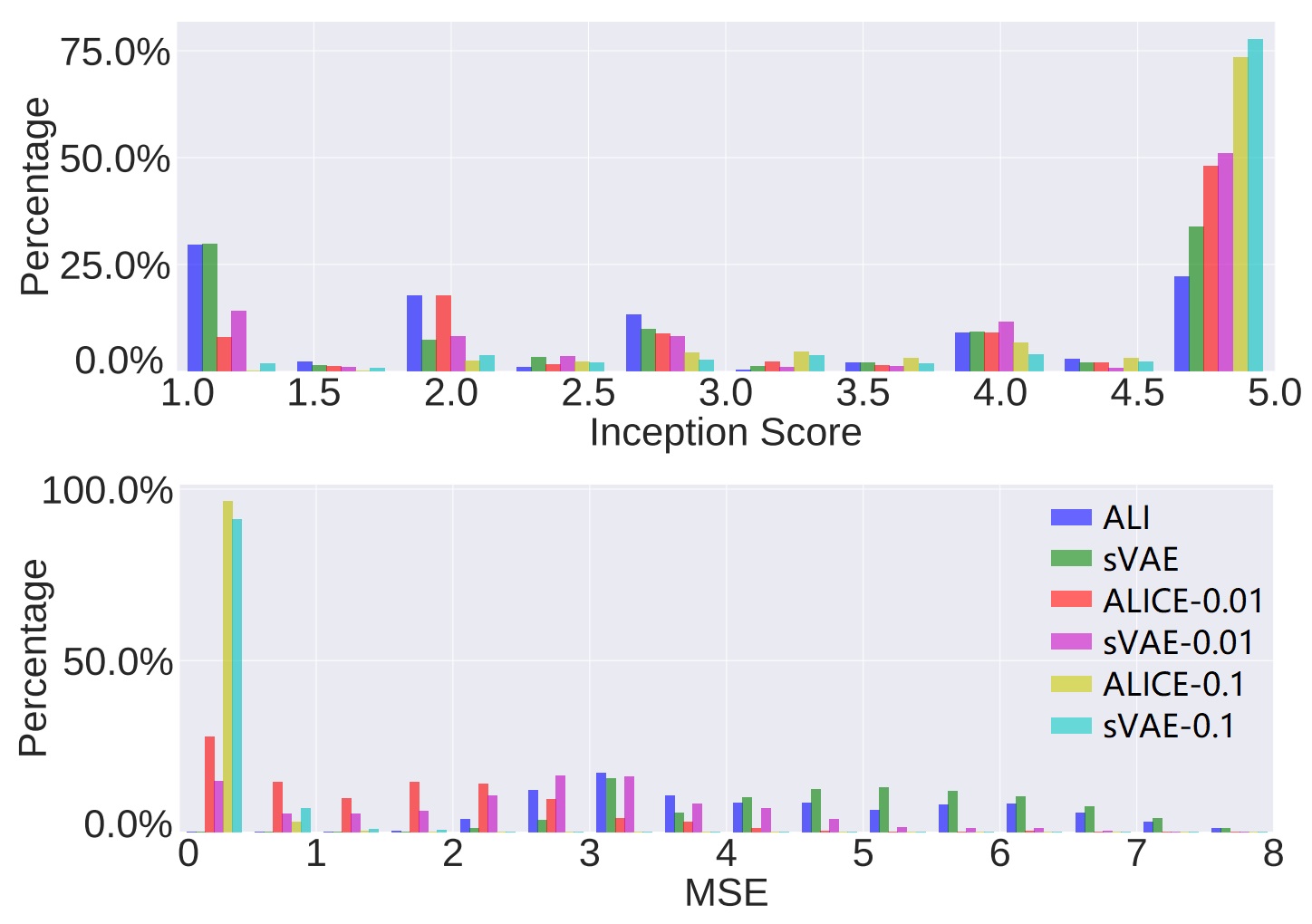}} 
		\caption{\small\label{fig:toydata} \text{sVAE results on toy dataset.} Top: Inception Score for ALI and sVAE with $\lambda = 0, 0.01, 0.1$. Bottom: Mean Squared Error (MSE).}
		\vspace{-4mm}
	\end{figure} 
	
\subsection{Toy Data}
In order to show the robustness and stability of our model, we test sVAE and sVAE-r on a toy dataset designed in the same manner as the one in ALICE \citep{lialice}. In this dataset, the true distribution of data $\xv$ is a two-dimensional Gaussian mixture model with five components. The latent code $\zv$ is a standard Gaussian distribution $\mathcal{N}(0,1)$. To perform the test, we consider using different values of $\lambda$ for both sVAE-r and ALICE. For each $\lambda$, $576$ experiments with different choices of architecture and hyper-parameters are conducted. In all experiments, we use mean square error (MSE) and inception score (IS) to evaluate the performance of the two models. Figure \ref{fig:toydata} shows the histogram results for each model. As we can see, both ALICE and sVAE-r are able to reconstruct images when $\lambda = 0.1$, while sVAE-r provides better overall inception score. 
	
	\begin{figure}[t]
		\centering 
		{ 
			\includegraphics[width=0.48\textwidth]{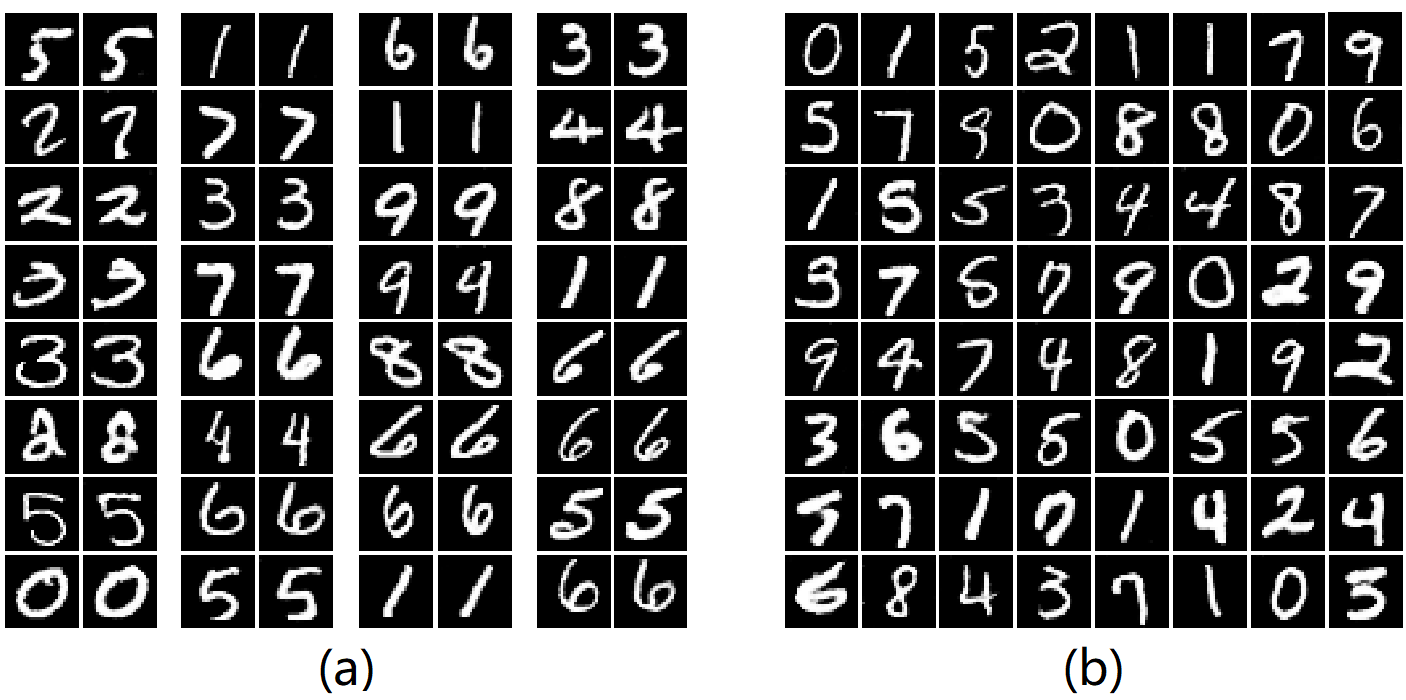}} 
		\caption{\small\label{fig:sVAE_mnist} \text{sVAE results on MNIST.} (a) and (b) are generated sample images by sVAE and sVAE-r, respectively. (c) is reconstructed images by sVAE-r: in each block, column one is ground-truth and column two is reconstructed images. Note that $\lambda$ is set to $0.1$ for sVAE-r.
		\vspace{-4mm}
		}
	\end{figure}  
	
\subsection{MNIST}
The results of image generation and reconstruction for sVAE, as applied to the MNIST dataset, are shown in Figure \ref{fig:sVAE_mnist}. By adding the regularization term, sVAE overcomes the limitation of image reconstruction in ALI. The log-likelihood of sVAE shown in Table \ref{tb:IS-mnist} is calculated using the annealed importance sampling method on the binarized MNIST dataset, as proposed in \citep{wu2016quantitative}. Note that in order to compare the model performance on binarized data, the output of the decoder is considered as a Bernoulli distribution instead of the Gaussian approach from the original paper. Our model achieves -79.26 nats, outperforming normalizing flow (-85.1 nats) while also being competitive to the state-of-the-art result (-79.2 nats). In addition, sVAE is able to provide compelling generated images, outperforming GAN \citep{gan} and WGAN-GP \citep{wgan-gp} based on the inception scores. 
	
	\begin{table}[h]
		\vspace{-4mm}
		\small
		\caption{\small Quantitative Results on MNIST. $\dagger$ is calculated using AIS. $\ddagger$ is reported in \citep{iwgan}.\label{tb:IS-mnist}}
		\begin{center}
			\begin{tabular}{lcc}
				\toprule
				{\bf Model} &$\text{log }p(x) \geq$ & {IS} \\
				\midrule
				NF (k=80)~\citep{normflows}         & -85.1       & - \\
				PixelRNN~\citep{pixelrnn}           & -79.2       & - \\
				AVB~\citep{AVB}                     & -79.5       & - \\
				ASVAE~\citep{Yunchen_NIPS17}       & -81.14      & - \\ 
				GAN~\citep{gan}          & -114.25 $^\dagger$           & 8.34  $^\ddagger$\\  
				WGAN-GP~\citep{wgan-gp}  & -79.92 $^\dagger$           & 8.45  $^\ddagger$\\  
				DCGAN~\citep{dcgan}       & -79.47 $^\dagger$       & 8.93 \\
				sVAE (ours)        & -80.42 $^\dagger$ & 8.81 \\ 
				sVAE-r (ours)      & -79.26 $^\dagger$ & 9.12 \\ 
				\bottomrule
			\end{tabular}
		\end{center}
	\end{table}	
	
	\begin{figure}[!tb]
		\centering{
			\includegraphics[width=0.425\textwidth]{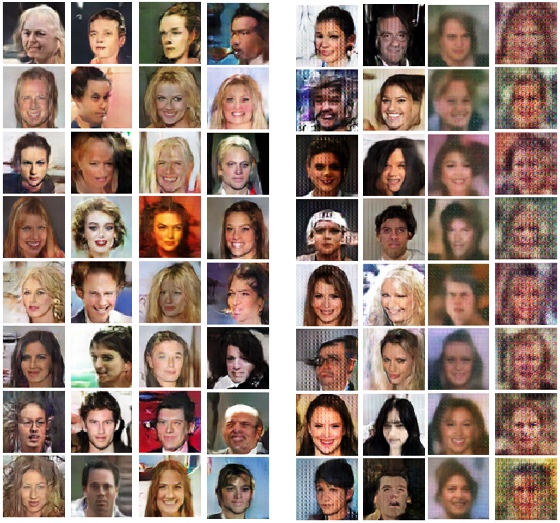}
			\caption{\small\label{fig:gen_face}\text{CelebA generation results.} Left block: sVAE-r generation. Right block: ALICE generation. $\lambda=0,0.1,1$ and $10$ from left to right in each block.} 
			\vspace*{\floatsep}
			\includegraphics[width=0.45\textwidth]{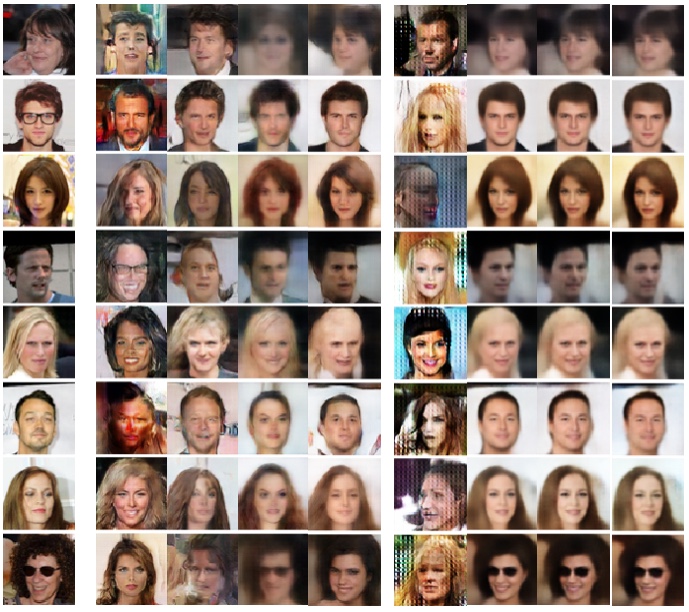}
			\caption{\small\label{fig:recon_face}\text{CelebA reconstruction results.} Left column: The ground truth. Middle block: sVAE-r reconstruction. Right block: ALICE reconstruction. $\lambda=0,0.1,1$ and $10$ from left to right in each block.}
			\vspace{-4mm}
		}
	\end{figure}
	
	\subsection{CelebA \label{sec:celebA}}
We evaluate sVAE on the CelebA dataset and compare the results with ALI. In experiments we note that for high-dimensional data like the CelebA, ALICE \citep{lialice} shows a trade-off between reconstruction and generation, while sVAE-r does not have this issue. If the regularization term is not included in ALI, the reconstructed images do not match the original images. On the other hand, when the regularization term is added, ALI is capable of reconstructing images but the generated images are flawed. In comparison, sVAE-r does well in both generation and reconstruction with different values of $\lambda$. The results for both sVAE and ALI are shown in Figure \ref{fig:gen_face} and \ref{fig:recon_face}.

	
	Generally speaking, adding the augmentation term as shown in (\ref{eq:augment}) should encourage more peaked $q_{\phiv}(\zv|\xv)$ and $p_{\thetav}(\xv|\zv)$. Nevertheless, ALICE fails in the inference process and performs more like an autoencoder. This is due to the fact that the discriminator becomes too sensitive to the regularization term. On the other hand, by using the symmetric KL (\ref{eq:newALI}) as the cost function, we are able to alleviate this issue, which makes sVAE-r a more stable model than ALICE. This is because sVAE updates the generator using the discriminator output, before the sigmoid, a non-linear transformation on the discriminator output scale.

	
	
	
	%
	
	\begin{figure}[!tb]
		\centering
		{
			\includegraphics[width=0.48\textwidth, height=3.5cm]{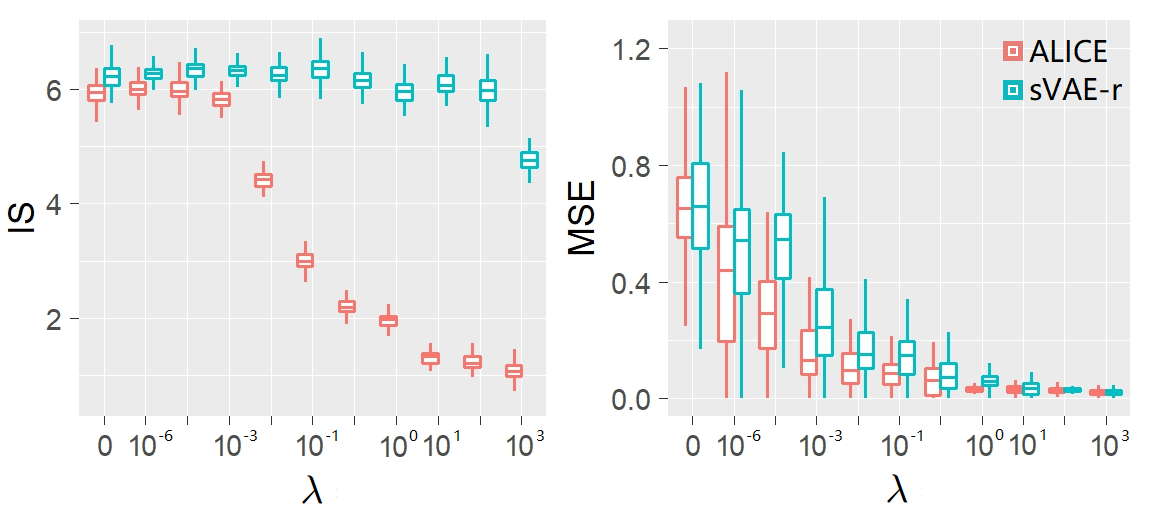}
			\caption{\small\label{fig:cifar_result} sVAE-r and ALICE CIFAR quantitative evaluation with different values of $\lambda$. Left: IS for generation; Right: MSE for reconstruction. The result is the average of multiple tests.}
			\vspace{-4mm}
		}
	\end{figure}

	\subsection{CIFAR-10}
	The trade-off of ALICE \citep{lialice} mentioned in Sec. \ref{sec:celebA} is also manifested in the results for the CIFAR-10 dataset. In Figure \ref{fig:cifar_result}, we show quantitative results in terms of inception score and mean squared error of sVAE-r and ALICE with different values of $\lambda$. As can be seen, both models are able to reconstruct images when $\lambda$ increases. However, when $\lambda$ is larger than $10^{-3}$, we observe a decrease in the inception score of ALICE, in which the model fails to generate images.
	
	\begin{table}[h]
		\small
		\caption{\small Unsupervised Inception Score on CIFAR-10} \label{tb:IS-cifar}
		\vspace{-2mm}
		\begin{center}
			\begin{tabular}{lcc}
				\toprule
				{\bf Model}  &{IS} \\
				\midrule
				ALI~\citep{ALI}                  & 5.34 $\pm$ .05 \\
				DCGAN~\citep{dcgan}              & 6.16 $\pm$ .07 \\
				ASVAE~\citep{Yunchen_NIPS17}   & 6.89 $\pm$ .05 \\
				WGAN-GP                          & 6.56 $\pm$ .05 \\
				WGAN-GP ResNet~\citep{wgan-gp}   & 7.86 $\pm$ .07 \\
				sVAE (ours)               & 6.76 $\pm$ .046 \\
				sVAE-r (ours)            & 6.96 $\pm$ .066 \\
				\bottomrule
			\end{tabular}
		\end{center}
	\end{table}	
	
	The CIFAR-10 dataset is also used to evaluate the generation ability of our model. The quantitative results, i.e., the inception scores, are listed in Table \ref{tb:IS-cifar}. Our model shows improved performance on image generation compared to ALI and DCGAN. Note that sVAE also gets comparable result as WGAN-GP \citep{wgan-gp} achieves. This can be interpreted using the similarity between (\ref{eq:newWGAN}) and (\ref{eq:wgan}) as summarized in the Sec. \ref{sec:previouswork}. The generated images are shown in Figure \ref{fig:svae_cifar_all}. More results are in the Appendix.

	{
		\begin{figure}[!tb]
		    \centering
			\begin{subfigure}[b]{0.47\textwidth}
			    \includegraphics[width=1\textwidth]{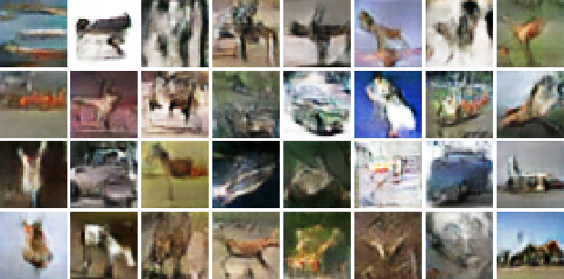}
			    \caption{sVAE CIFAR unsupervised generation.}
			    \label{fig:svae_cifar_a}
			    \vspace{2mm}
			\end{subfigure}
			\begin{subfigure}[b]{0.47\textwidth}
			    \includegraphics[width=1\textwidth]{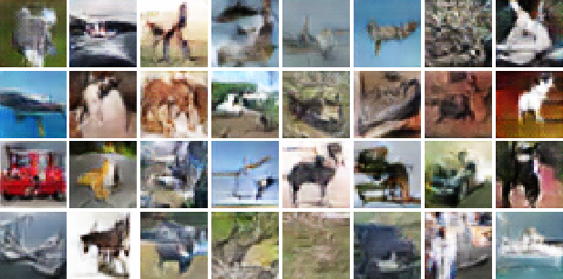}
			    \caption{sVAE-r (with $\lambda=1$) CIFAR unsupervised generation.}
			    \label{fig:svae_cifar_b}
			    \vspace{2mm}
			\end{subfigure}
			\begin{subfigure}[b]{0.47\textwidth}
			    \includegraphics[width=1\textwidth]{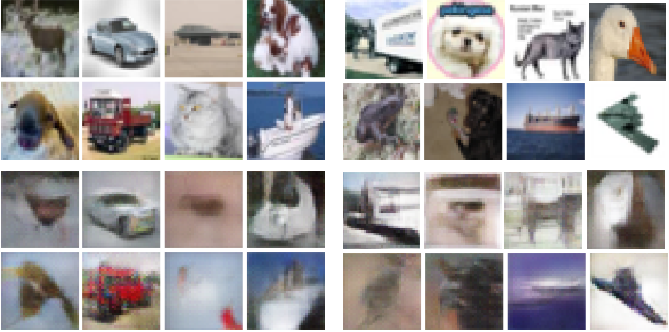}
			    \caption{sVAE-r (with $\lambda=1$) CIFAR unsupervised reconstruction. First two rows are original images, and the last two rows are the reconstructions.}
			    \label{fig:svae_cifar_c}
			\end{subfigure}
			\caption{sVAE CIFAR results on image generation and reconstruction.}
			\label{fig:svae_cifar_all}
			\vspace{-2mm}
		\end{figure}
		
		\section{Conclusions}
		We present the symmetric variational autoencoder (sVAE), a novel framework which can match the joint distribution of data and latent code using the {\em symmetric} Kullback-Leibler divergence. The experiment results show the advantages of sVAE, in which it not only overcomes the missing mode problem \citep{iwgan}, but also is very stable to train. With excellent performance in image generation and reconstruction, we will apply sVAE on semi-supervised learning tasks and conditional generation tasks in future work. Morever, because the latent code $z$ can be treated as data from a different domain, \ie, images \citep{cycle, discogan} or text \citep{gan2017triangle}, we can also apply sVAE to domain transfer tasks.

		\newpage
		{
			{
				\bibliographystyle{abbrvnat}
				\bibliography{LRSref,nips2017}}
		}
		
		 Appendix
 		\clearpage
 		\newpage
 		\appendix
 		\onecolumn
		
 		\section{Model Architectures}
 		\begin{table}[h!]
 			\caption{\small Architecture of the models for sVAE-r on MNIST. BN denotes batch normalization.}
 			\vskip 0.05in
 			\hspace{-10mm}
 			\small
 			\begin{tabular}{c|c|c}
 				\toprule
 				Encoder X to z  & Decoder z to X  & Discriminator \\
 				\midrule
 				Input $28\times28$ Gray Image & Input latent code z & Input two $28\times28$ Gray Image\\
				
 				\midrule
 				$5\times5$ conv. 16 ReLU, stride 2, BN & MLP output 1024, BN & $5\times5$ conv. 32 ReLU, stride 2, BN\\
 				$5\times5$ conv. 32 ReLU, stride 2, BN & MLP output 3136, BN & $5\times5$ conv. 64 ReLU, stride 2, BN\\
 				MLP output 784, BN & $5\times5$ deconv. 64 ReLU, stride 2, BN & $5\times5$ conv. 128 ReLU, stride 2, BN\\
 				& &input z through MLP output 1024, ReLU \\
 				MLP output dim of z & $5\times5$ deconv. 1 ReLU, stride 2, sigmoid & MLP output 1 \\
				
 				\bottomrule
 			\end{tabular} 
 			\label{Table:sVAE_model_mnist}
 		\end{table}
		
 		\begin{table}[h!]
 			\caption{\small Architecture of the models for sVAE on CelebA. BN denotes batch normalization. lReLU denotes Leaky ReLU.}
 			\vskip 0.05in
 			\hspace{-10mm}
 			\small
 			\begin{tabular}{c|c|c}
 				\toprule
 				Encoder X to z  & Decoder z to X  & Discriminator \\
 				\midrule
 				Input Image X concat with noise & Input z concat with noise & Input X\\
				
 				\midrule
 				$4\times4$ conv. 32 lReLU, stride 2, BN & concat random noise & $5\times5$ conv. 64 ReLU, stride 2, BN\\
 				$4\times4$  conv. 64 lReLU, stride 2, BN & MLP output 1024, lReLU, BN & $5\times5$ conv. 128 ReLU, stride 2, BN\\
 				$4\times4$ conv. 128 lReLU, stride 2, BN & MLP output 8192, lReLU, BN & $5\times5$ conv. 256 ReLU, stride 2, BN\\
 				$4\times4$ conv. 256 lReLU, stride 2, BN &  & $5\times5$ conv. 512 ReLU, stride 2, BN\\
 				$4\times4$ conv. 512 lReLU, stride 2, BN & $5\times5$ deconv. 256 lReLU, stride 2, BN  & Input z through MLP, output 2046, ReLU\\
 				MLP output 512, lReLU  & $5\times5$ deconv. 128 lReLU, stride 2, BN  & concat two features from X and z\\
 				MLP output dim of z, tanh & $5\times5$ deconv. 64 lReLU, stride 2, BN  & \\
 				& $5\times5$ deconv. 3 tanh, stride 2, BN  & MLP output 1 \\
				
 				\bottomrule
 			\end{tabular} 
 			\label{Table:sVAE_model_celebA}
 		\end{table}
		
 		\begin{table}[h!]
 			\caption{\small Architecture of the models for sVAE-r on CIFAR. BN denotes batch normalization. lReLU denotes Leaky ReLU. $Dim$ denotes the number of attributes.}
 			\vskip 0.05in
 			\hspace{-10mm}
 			\small
 			\begin{tabular}{c|c|c}
 				\toprule
 				Encoder X to z  & Decoder z to X  & Discriminator \\
 				\midrule
 				Input Image X concat with noise & Input z & Input X  \\
				
 				\midrule
 				$5\times5$ conv. 32 lReLU, stride 2, BN & concat random noise & $5\times5$ conv. 64 ReLU, stride 2, BN\\
 				$5\times5$  conv. 64 lReLU, stride 2, BN &  & $5\times5$ conv. 128 ReLU, stride 2, BN\\
 				$5\times5$ conv. 128 lReLU, stride 2, BN & MP output 8192, lReLU, BN & $5\times5$ conv. 256 ReLU, stride 2, BN\\
 				$5\times5$ conv. 256 lReLU, stride 2, BN & & $5\times5$ conv. 512 ReLU, stride 2, BN, avg pooling\\
 				& $5\times5$ deconv. 256 ReLU, stride 2, BN  & Input z through MLP, output 512, ReLU\\
 				MLP output 512, lReLU  & $5\times5$ deconv. 128 ReLU, stride 2, BN  & concat two features from X and z\\
 				MLP output dim of z, tanh & $5\times5$ deconv. 3 tanh, stride 2  & MLP output 1\\
				
 				\bottomrule
 			\end{tabular} 
 			\label{Table:sVAE_model_cifar}
 		\end{table}
		
 		\section{More Result}
 		\subsection{CIFAR-10 result}
 		\begin{figure}[h!]
 			\centering
 			\includegraphics[width=\textwidth]{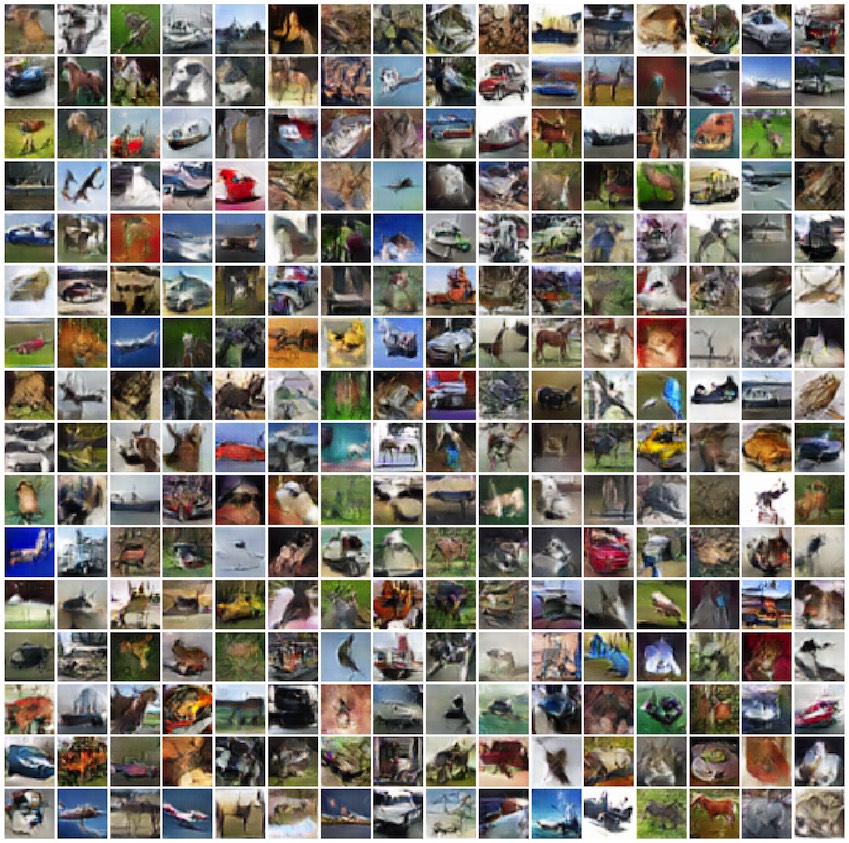}
 			\caption{\small\label{fig:svae_more_cifar}\text{sVAE CIFAR unsupervised generation results with $\lambda=0.1$.}}
 		\end{figure}

 		\begin{figure}[h!]
 			\centering
 			{
 				\includegraphics[width=\textwidth]{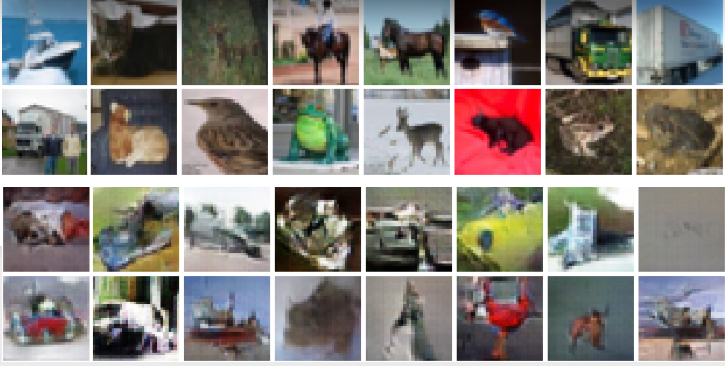}
 				\caption{\small\label{fig:svae_cifar_recon} sVAE CIFAR unsupervised reconstruction. First two rows are original images, and the last two rows are the reconstructions  }
 			}
 		\end{figure}
		
 		\begin{figure}[h!]
 			\centering
 			{
 				\includegraphics[width=\textwidth]{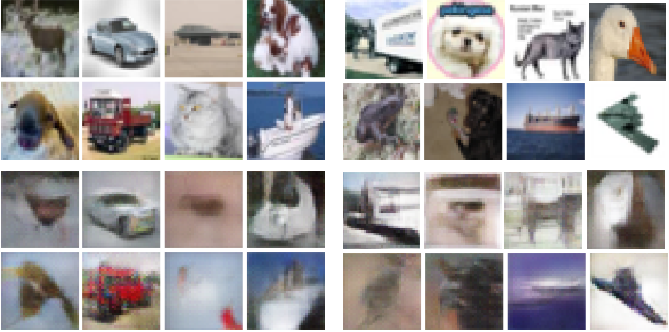}
 				\caption{\small\label{fig:svae-r_cifar_recon} sVAE-r CIFAR unsupervised reconstruction. First two rows are original images, and the last two rows are the reconstructions  }
 			}
 		\end{figure}
		
 		\subsection{CelebA result}
 		\begin{figure}
 			\centering
 			{
 				\includegraphics[width=\textwidth]{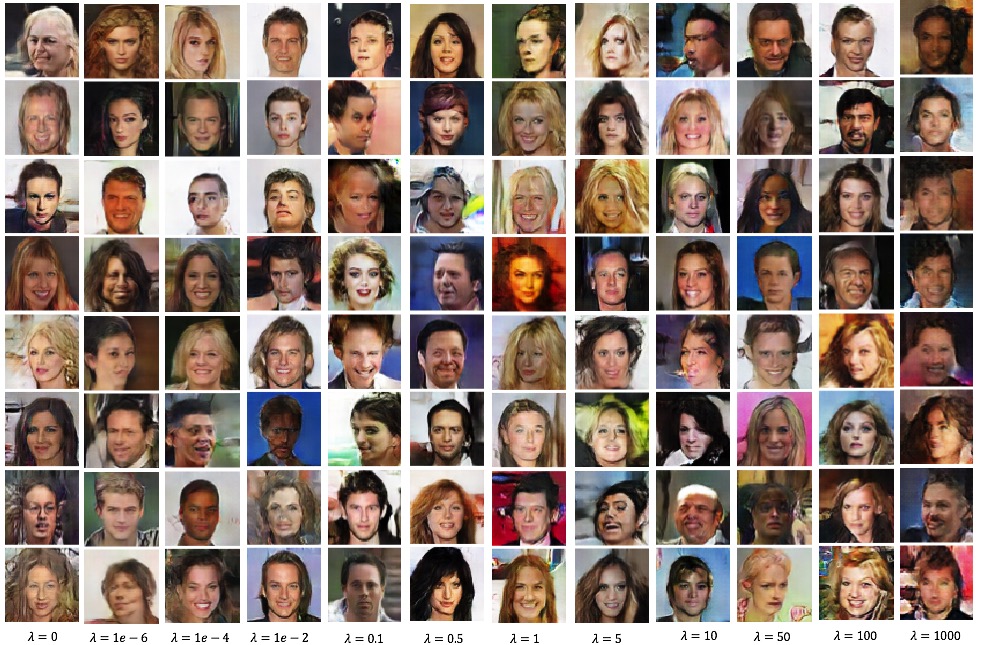}
 				\caption{\small\label{fig:svae_face}\text{sVAE-r CelebA generations results with different $\lambda$}} 
 			}
 		\end{figure}
		
 		\begin{figure}
 			\centering
 			{
 				\includegraphics[width=\textwidth]{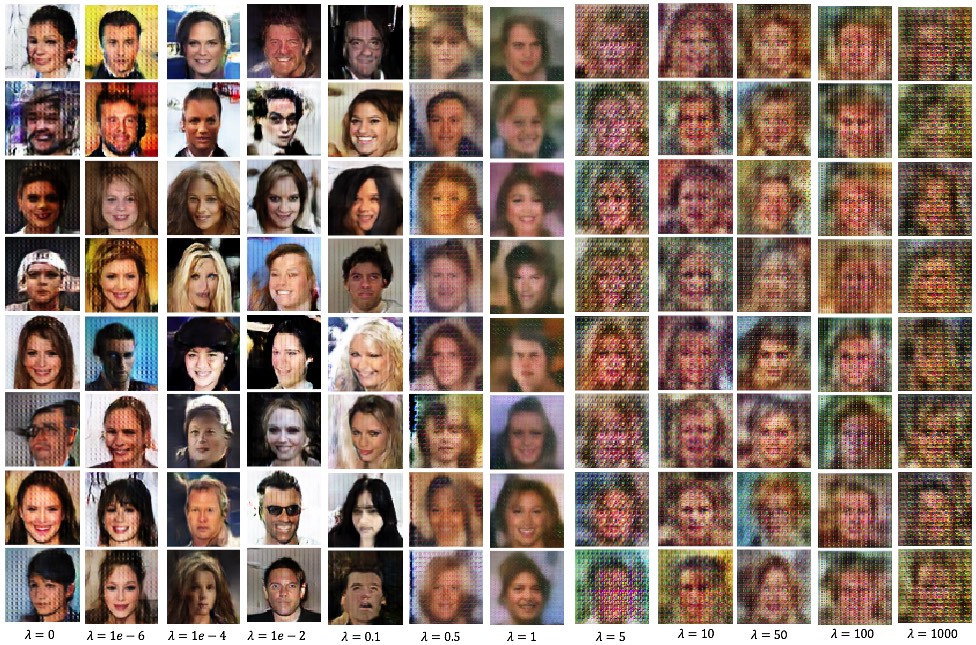}
 				\caption{\small\label{fig:svae-r_face}\text{ALICE CelebA generations results with different $\lambda$} } 
 			}
 		\end{figure}
		
 		\begin{figure}
 			\centering
 			{
 				\includegraphics[width=\textwidth]{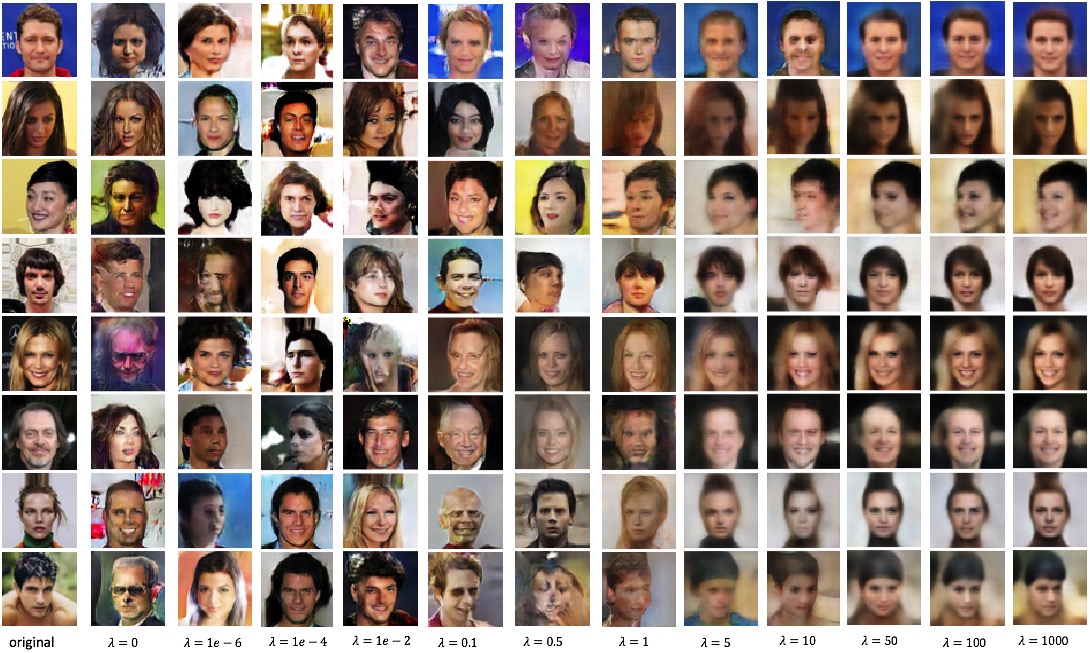}
 				\caption{\small\label{fig:ali_face}\text{ALICE CelebA reconstructions with different $\lambda$.} }
 			}
 		\end{figure}
		
 		\begin{figure}
 			\centering
 			{
 				\includegraphics[width=\textwidth]{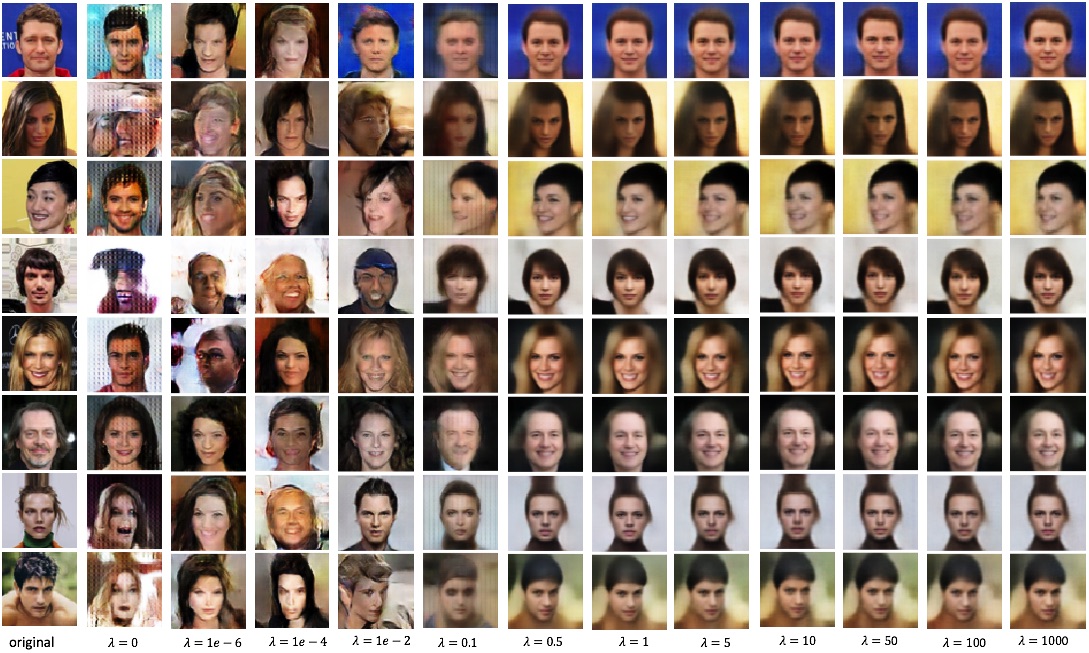}
 				\caption{\small\label{fig:ALI-r_face}\text{ALICE CelebA reconstructions with different $\lambda$.} }
 			}
 		\end{figure}

	\end{document}